\documentclass[journal]{IEEEtran}
\usepackage{cite}
\usepackage{amsmath,amssymb,amsfonts}
\usepackage{algorithmic}
\usepackage{graphicx}
\usepackage{algorithm,algorithmic}
\usepackage{hyperref}
\hypersetup{hidelinks=true}
\usepackage{textcomp}

\usepackage{booktabs}
\usepackage{multirow}
\usepackage{makecell}
\usepackage{pifont}
\usepackage{subcaption}

\usepackage{orcidlink}
\usepackage{pifont}

\ifCLASSINFOpdf

\else

\fi

\begin{document}

\title{Privileged Lesion-Context Relational Distillation for Mask-Free Skin Lesion Classification}

\author{
Abu Mukaddim Rahi \orcidlink{0009-0002-8875-4511},
Md Mithun Hossain \orcidlink{0009-0001-4883-1802},
Md Zulficar Hasan Joy \orcidlink{0009-0001-4151-6818},
and M. F. Mridha \orcidlink{0000-0001-5738-1631}, ~\IEEEmembership{Senior Member,~IEEE},
Md. Jakir Hossen \orcidlink{0000-0002-9978-7987}
\thanks{Manuscript received [Month Day, Year]; revised [Month Day, Year]; accepted [Month Day, Year]. This research received no external funding.}%
\thanks{Abu Mukaddim Rahi is with the Department of IT in Artificial Intelligence, Victoria University, Footscray, VIC, Australia (e-mail: mukaddimrahi@gmail.com).}
\thanks{Md Mithun Hossain is with the BUBT Resarch Graduate School, Bangaldesh University of Business and Technology, Dhaka, Bangladesh (e-mail: mithunhossain@bubt.edu.bd).}
\thanks{Md Zulficar Hasan Joy is with the Department of Artificial Intelligence, Daffodil Institute of Information Technology, Dhaka, Bangladesh (e-mail: zulficarjoy247@gmail.com).}%
\thanks{M. F. Mridha is with the Department of Computer Science and Engineering, American International University-Bangladesh (AIUB), Dhaka 1229, Bangladesh (e-mail: firoz.mridha@aiub.edu).}%
\thanks{Md. Jakir Hossen is with the Center for Advanced Analytics, COE for Artificial Intelligence, Faculty of Engineering and Technology, Multimedia University, Melaka 75450, Malaysia (e-mail: jakir.hossen@mmu.edu.my).}%
\thanks{Corresponding author: Md. Jakir Hossen (e-mail: jakir.hossen@mmu.edu.my).}
}

\markboth{Journal of \LaTeX\ Class Files,~Vol.~14, No.~8, August~2015}%
{A. M. Rahi \& M. M. Hossain \MakeLowercase{\textit{et al.}}: Privileged Lesion-Context Relational Distillation for Mask-Free Skin Lesion Classification}

\maketitle

\begin{abstract}
Accurate skin lesion classification can benefit from lesion segmentation masks, but requiring masks or an auxiliary segmentation model during inference reduces clinical practicality and increases computational complexity. This work introduces Privileged Lesion-Context Relational Distillation (PLCRD), a teacher-student framework that exploits lesion masks exclusively during training while preserving image-only inference. The privileged teacher jointly analyzes the original dermoscopic image and its mask-guided lesion region to learn lesion-specific and contextual diagnostic representations. An image-only student is then trained through complementary knowledge-transfer mechanisms that convey the teacher’s diagnostic distribution, lesion-focused attention, inter-lesion relational geometry, and lesion-context structure. PLCRD decomposes deep representations into lesion and contextual embeddings and transfers their relational organization through inter-lesion similarity alignment, lesion-context affinity matching, separation regularization, and class-aware relational learning. This formulation avoids direct feature matching between heterogeneous teacher and student architectures and enables the student to internalize mask-informed diagnostic structure without accessing masks at deployment. The framework was evaluated on HAM10000 using lesion-disjoint data partitioning and externally validated on ISIC 2018 without retraining. PLCRD achieved a lesion-level macro-F1 of \textbf{0.773 $\pm$ 0.018}, balanced accuracy of \textbf{0.764 $\pm$ 0.023}, and macro-AUROC of \textbf{0.976 $\pm$ 0.002} on HAM10000, together with a macro-F1 of \textbf{0.732 $\pm$ 0.008} on ISIC 2018. The results indicate that privileged lesion annotations can be transformed into transferable relational knowledge, yielding a practical and interpretable approach to mask-free skin lesion classification.
\end{abstract}


%
\IEEEpeerreviewmaketitle

\section{Introduction}
\label{sec:introduction}

\IEEEPARstart{S}{kin} lesion classification from dermoscopic images has received considerable attention because early and accurate recognition of malignant lesions can support timely clinical intervention. Deep learning models have demonstrated strong diagnostic performance and, under controlled settings, have achieved results comparable to dermatologists~\cite{esteva2017dermatologist,haenssle2018man,tschandl2019comparison}. Public datasets such as HAM10000 and the International Skin Imaging Collaboration archives have further accelerated the development of automated dermoscopic image analysis systems~\cite{tschandl2018ham10000,codella2019skin}.
Despite this progress, reliable skin lesion classification remains challenging. Benign and malignant lesions may share similar visual characteristics, while images exhibit substantial variation in pigmentation, texture, illumination, acquisition devices, and surrounding skin. Deep classifiers may also exploit non-lesion cues, including rulers, markings, image borders, and dataset-specific artifacts, rather than clinically meaningful lesion structures. Such behavior can reduce interpretability and limit generalization to external data collected under different conditions.

Lesion segmentation masks can provide explicit spatial supervision and guide a classifier toward diagnostically relevant regions. However, requiring ground-truth masks or an auxiliary segmentation model during inference increases annotation requirements, computational complexity, and sensitivity to segmentation errors. Moreover, isolating the lesion may remove useful information from the surrounding skin, since lesion boundaries, local contrast, and contextual color patterns can contribute to diagnosis~\cite{mahbod2020effects}. A more practical strategy is therefore to exploit masks as privileged information during training while preserving image-only inference.
Learning using privileged information provides a suitable framework for this setting, as it allows additional supervisory information to be available during training but absent during deployment~\cite{vapnik2009new}. Knowledge distillation can transfer information from a privileged teacher to a deployable student through output distributions, intermediate representations, or attention patterns~\cite{hinton2015distilling,zagoruyko2017paying}. Nevertheless, conventional distillation methods primarily align logits or raw features and do not explicitly model the relational structure between lesion-specific evidence and surrounding context. Direct feature alignment is also restrictive when teacher and student networks have heterogeneous architectures and different feature dimensions.

To address these limitations, we propose Privileged Lesion-Context Relational Distillation (PLCRD), a teacher-student framework for mask-free skin lesion classification. During training, a privileged dual-branch teacher processes the original dermoscopic image together with a mask-guided lesion view, whereas the student receives only the original image. The teacher transfers diagnostic distributions and class-conditioned attention to the student. More importantly, PLCRD decomposes deep representations into lesion-focused and contextual embeddings and transfers their relational organization rather than directly matching raw feature vectors.
The proposed relational objective aligns inter-lesion similarity structures, preserves lesion-context affinity, prevents collapse between lesion and contextual representations, and encourages class-aware organization of lesion embeddings. Because the transfer is defined through similarities and affinities, PLCRD supports heterogeneous teacher and student architectures without requiring identical feature dimensions. At inference, the teacher and lesion masks are discarded, and only the image-based student is retained.

The main contributions of this work are summarized as follows:

\begin{itemize}
    \item We propose PLCRD, a privileged teacher-student framework that uses lesion masks only during training and requires only dermoscopic images during inference.

    \item We design a dual-branch privileged teacher that jointly captures global image context and explicitly localized lesion information.

    \item We introduce a lesion-context relational distillation objective that transfers inter-lesion geometry, lesion-context affinity, separation, and class-aware relational structure between heterogeneous networks.

    \item We integrate relational transfer with diagnostic-distribution and class-conditioned attention distillation to provide complementary semantic and spatial supervision.
\end{itemize}

The remainder of this paper is organized as follows. Section~\ref{sec:related_work} reviews the relevant literature. Section~\ref{sec:method} presents the proposed PLCRD framework. Section~\ref{sec:experiments} describes the experimental setup, and Section~\ref{sec:results} reports the results. Section~\ref{sec:limitations_future_work} discusses the limitations and future work, while Section~\ref{sec:conclusion} concludes the paper.

\section{Related Work}
\label{sec:related_work}

\subsection{Deep Learning for Dermoscopic Image Classification}
\label{subsec:deep_learning_dermoscopy}

Deep learning has substantially improved automated dermoscopic image classification by learning hierarchical visual representations directly from lesion images. In addition to early dermatologist-comparison studies, several works have evaluated deep models across broader diagnostic settings and lesion subtypes. Marchetti \textit{et al.} compared computer algorithms with dermatologists for melanoma recognition in the ISIC challenge setting, demonstrating the increasing competitiveness of automated systems for dermoscopic diagnosis~\cite{marchetti2018results}. Tschandl \textit{et al.} investigated convolutional neural networks for nonpigmented skin cancer diagnosis and reported expert-level performance under controlled evaluation conditions~\cite{tschandl2019nonpigmented}. Yu \textit{et al.} further explored convolutional networks for acral melanoma detection, highlighting the potential of deep models for difficult lesion subtypes~\cite{yu2018acral}. Public datasets and challenges have also shaped progress in this field. The ISIC challenge series provided standardized benchmarks for lesion segmentation and classification~\cite{gutman2016isic,codella2018isbi,codella2019skin}, while HAM10000 enabled large-scale multi-class evaluation of common pigmented skin lesions~\cite{tschandl2018ham10000}. To improve classification accuracy, prior studies have explored hybrid networks, multi-level ensembles, and stronger image encoders~\cite{mahbod2017hybrid,xie2018mlde}. Modern architectures such as EfficientNet, MobileNetV3, Swin Transformer, and mobile-scale vision transformers provide powerful and efficient backbones for image classification~\cite{tan2019efficientnet,howard2019searching,liu2021swin,li2023rethinking}. However, architectural improvements alone do not ensure that a model relies on clinically meaningful lesion evidence rather than background artifacts or dataset-specific shortcuts.

\subsection{Segmentation-Guided Classification}
\label{subsec:segmentation_guided_classification}

Lesion segmentation has commonly been used to guide dermoscopic classification by isolating or emphasizing the lesion region. Earlier segmentation-classification pipelines relied on lesion boundaries to reduce background influence and improve interpretability~\cite{sumithra2016segmentation}. In deep learning systems, masks may be used to crop lesions, suppress non-lesion regions, provide auxiliary supervision, or train joint segmentation-classification networks. Although lesion masks provide useful spatial information, segmentation-guided classification has important limitations. Mahbod \textit{et al.} showed that segmentation does not always improve dermoscopic image classification, indicating that removing surrounding context may discard useful diagnostic cues~\cite{mahbod2020effects}. Contextual information such as lesion-boundary transition, local contrast, erythema, and surrounding pigmentation can support diagnosis. Moreover, a test-time segmentation model introduces additional computational cost and can propagate segmentation errors into the classifier. These issues motivate methods that use masks as training-time guidance while preserving mask-free inference.

\subsection{Learning Using Privileged Information and Knowledge Distillation}
\label{subsec:lupi_kd}

Learning using privileged information allows a model to exploit additional information during training that is unavailable at inference~\cite{vapnik2009new}. In medical imaging, such information may include lesion masks, expert annotations, clinical metadata, or pathology-derived signals. For dermoscopic classification, segmentation masks are a natural privileged modality because they identify lesion location without needing to be available during deployment. Knowledge distillation provides a practical mechanism for transferring privileged knowledge from a teacher to an image-only student. Response-based distillation trains the student to match softened teacher predictions, allowing it to learn inter-class similarity and uncertainty~\cite{hinton2015distilling}. Feature-based distillation uses intermediate teacher representations as hints~\cite{romero2015fitnets}, while attention transfer encourages the student to reproduce spatial attention patterns from the teacher~\cite{zagoruyko2017paying}. These approaches are useful for compact or deployable models, but they remain limited in the present setting. Logit distillation transfers diagnostic distributions but not lesion-context structure; feature matching can be restrictive when teacher and student architectures have different feature dimensions; and attention transfer provides spatial guidance without explicitly modeling relationships among lesions or between lesion and context.

\subsection{Position of the Proposed Method}
\label{subsec:position_plcrd}

The proposed PLCRD framework differs from conventional segmentation-guided and distillation-based methods in three main ways. First, lesion masks are used only to train a privileged teacher, while the final student requires only the original dermoscopic image at inference. Second, PLCRD does not depend on direct feature matching between teacher and student networks. Instead, it transfers relational structures defined by similarities and affinities, making it suitable for heterogeneous teacher-student architectures. Third, PLCRD explicitly models lesion-context organization by decomposing representations into lesion-focused and contextual embeddings.

Through inter-lesion relation alignment, lesion-context affinity matching, separation regularization, and class-aware relational learning, PLCRD converts privileged mask supervision into deployable relational knowledge. This allows the student to benefit from mask-informed training while avoiding the annotation and computational burdens of mask-dependent inference.
\section{Proposed Method}
\label{sec:method}

\subsection{Problem Formulation}
\label{subsec:problem_formulation}

Let $\mathcal{D}=\{(\mathbf{x}_i,\mathbf{m}_i,y_i)\}_{i=1}^{N}$ denote the training dataset, where $\mathbf{x}_i \in \mathbb{R}^{3 \times H \times W}$ is a dermoscopic image, $\mathbf{m}_i \in \{0,1\}^{1 \times H \times W}$ is the corresponding lesion mask, and $y_i \in \{1,\ldots,C\}$ is the diagnostic class label. The objective is to learn a classifier that predicts the lesion category using only the image,

\begin{equation}
    \hat{y}_i = \arg\max_c p_{\theta}(y=c \mid \mathbf{x}_i),
\end{equation}

without requiring a lesion mask during inference. In this setting, the mask is treated as privileged information: it is available during training to guide representation learning but is not used during deployment.

\subsection{Overview of PLCRD}
\label{subsec:overview_plcrd}

\begin{figure*}[t]
    \centering
    \includegraphics[width=\textwidth]{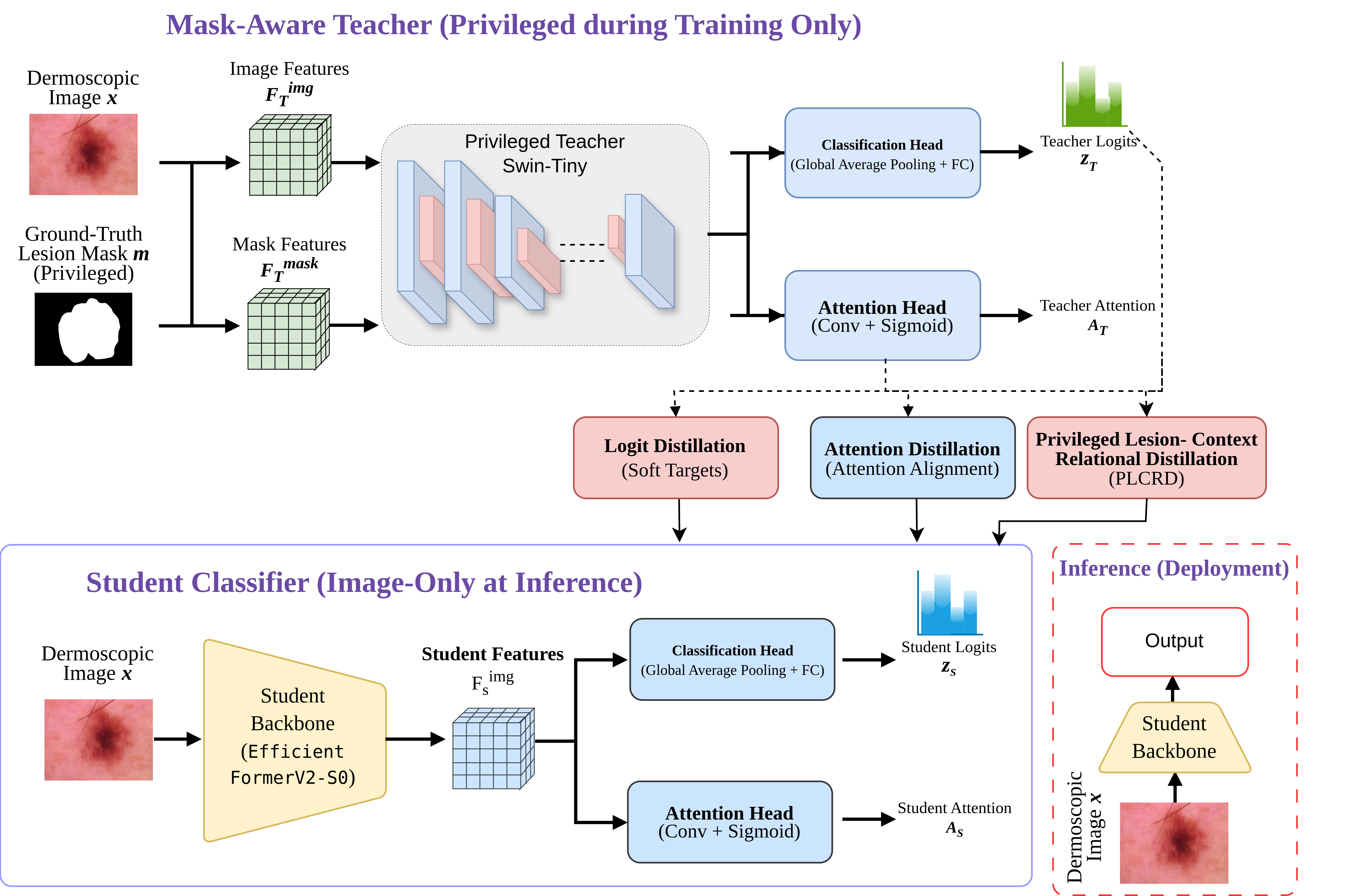}
    \caption{Overview of the proposed PLCRD framework. The mask-aware teacher uses lesion masks as privileged information during training, while the student classifier receives only the original dermoscopic image. Knowledge is transferred through logit distillation, class-conditioned attention distillation, and privileged lesion-context relational distillation. During inference, only the image-only student is retained.}
    \label{fig:framework}
\end{figure*}

Fig.~\ref{fig:framework} presents the proposed Privileged Lesion-Context Relational Distillation (PLCRD) framework. PLCRD consists of a mask-aware teacher and an image-only student. During training, the teacher receives both the original dermoscopic image and a mask-guided lesion view. This allows the teacher to learn diagnostic representations that combine global image context with explicitly localized lesion information. The student receives only the original image and is therefore suitable for mask-free deployment.
The teacher transfers knowledge to the student through three complementary signals: softened diagnostic predictions, class-conditioned attention maps, and lesion-context relational structure. The softened predictions transfer inter-class diagnostic uncertainty, the attention maps transfer spatial evidence, and the relational objective transfers how lesion-focused and contextual representations are organized. Unlike direct feature matching, PLCRD transfers relationships among embeddings. Therefore, it can be applied even when the teacher and student use different backbone architectures or feature dimensions.

\subsection{Privileged Mask-Aware Teacher}
\label{subsec:privileged_teacher}

The privileged teacher is designed to exploit lesion masks only during training. Given an image $\mathbf{x}_i$ and its mask $\mathbf{m}_i$, a lesion-focused image is generated as

\begin{equation}
    \mathbf{x}^{l}_i = \mathbf{x}_i \odot \mathbf{m}_i,
\end{equation}

where $\odot$ denotes element-wise multiplication. The teacher processes the original image and the lesion-focused image through two complementary branches. The original-image branch captures global contextual information, while the lesion branch emphasizes localized lesion morphology. Let $\mathbf{F}^{T,c}_i$ and $\mathbf{F}^{T,l}_i$ denote the context and lesion feature maps, respectively. These features are fused to form the final teacher representation,

\begin{equation}
    \mathbf{F}^{T}_i =
    \psi_T
    \left(
    [\mathbf{F}^{T,c}_i;\mathbf{F}^{T,l}_i]
    \right),
\end{equation}

where $[\cdot;\cdot]$ denotes channel-wise concatenation and $\psi_T(\cdot)$ is a learnable fusion block. The fused feature map is passed through global average pooling and a classification head to produce teacher logits $\mathbf{o}^{T}_i$ and probabilities $\mathbf{p}^{T}_i$.

The teacher is optimized using classification supervision and mask-aware localization guidance. The localization term encourages attention to remain concentrated inside the lesion while suppressing excessive attention outside the lesion. Let $\mathbf{A}^{T}_i$ be the teacher attention map and $\widetilde{\mathbf{m}}_i$ be the lesion mask resized to the attention resolution. The inside- and outside-lesion attention ratios are computed as

\begin{equation}
    r_i^{\mathrm{in}} =
    \frac{\sum_{u,v} A_i^{T}(u,v)\widetilde{m}_i(u,v)}
    {\sum_{u,v}\widetilde{m}_i(u,v)+\epsilon},
    \quad
    r_i^{\mathrm{out}} =
    \frac{\sum_{u,v} A_i^{T}(u,v)(1-\widetilde{m}_i(u,v))}
    {\sum_{u,v}(1-\widetilde{m}_i(u,v))+\epsilon}.
\end{equation}

The teacher is encouraged to increase lesion-region attention and reduce non-lesion attention through a weak localization loss. After teacher training, the teacher parameters are frozen and used only to supervise the student.

\subsection{Image-Only Student}
\label{subsec:image_only_student}

The student receives only the original dermoscopic image. Given $\mathbf{x}_i$, the student backbone extracts a feature map $\mathbf{F}^{S}_i$, which is followed by global average pooling and a classification head to produce logits $\mathbf{o}^{S}_i$ and class probabilities $\mathbf{p}^{S}_i$. The supervised classification loss is

\begin{equation}
    \mathcal{L}_{\mathrm{cls}}
    =
    -\frac{1}{N}
    \sum_{i=1}^{N}
    w_{y_i}\log p^{S}_{i,y_i},
\end{equation}

where $w_{y_i}$ is a class-balancing weight. This weighting reduces the effect of class imbalance, which is important for dermoscopic datasets where the majority class can dominate the training objective.

\subsection{Logit and Attention Distillation}
\label{subsec:logit_attention_distillation}

The first distillation signal transfers the teacher's diagnostic distribution to the student. Given temperature $\tau$, the logit distillation loss is

\begin{equation}
    \mathcal{L}_{\mathrm{KD}}
    =
    \tau^{2}
    D_{\mathrm{KL}}
    \left(
    \operatorname{softmax}
    \left(
    \frac{\mathbf{o}^{T}}{\tau}
    \right)
    \middle\|
    \operatorname{softmax}
    \left(
    \frac{\mathbf{o}^{S}}{\tau}
    \right)
    \right),
\end{equation}

where $D_{\mathrm{KL}}(\cdot\|\cdot)$ denotes the Kullback--Leibler divergence. This loss transfers inter-class similarity and diagnostic uncertainty from the privileged teacher to the image-only student.

The second distillation signal transfers spatial evidence. Both teacher and student generate class-conditioned attention maps from their final feature maps. For model $q \in \{T,S\}$, the attention map is computed as

\begin{equation}
    \mathbf{A}^{q}
    =
    \sum_{c=1}^{C}
    p^{q}_{c}
    \sigma
    \left(
    \operatorname{Conv}_{1\times1}^{q}(\mathbf{F}^{q})_c
    \right),
\end{equation}

where $\sigma(\cdot)$ denotes the sigmoid function. This formulation weights class-specific evidence maps by the predicted class probabilities, allowing attention to reflect diagnostic uncertainty. The teacher attention map is resized to the student resolution and used as a soft spatial target:

\begin{equation}
    \mathcal{L}_{\mathrm{att}}
    =
    \operatorname{SmoothL1}
    \left(
    \mathbf{A}^{S},
    \widetilde{\mathbf{A}}^{T}
    \right).
\end{equation}

This encourages the student to focus on spatial regions emphasized by the mask-aware teacher while still requiring only the image at inference.

\subsection{Privileged Lesion-Context Relational Distillation}
\label{subsec:plcrd}

The central component of PLCRD is relational transfer between lesion-focused and contextual representations. Instead of forcing the student to reproduce the teacher's raw feature vectors, PLCRD transfers the relational structure learned by the privileged teacher. This is important because the teacher and student may have different feature dimensions and representational capacities.
For each model $q \in \{T,S\}$, the attention map is used to decompose the feature map into a lesion embedding and a contextual embedding. The lesion embedding is obtained through attention-weighted pooling, and the contextual embedding is obtained using the complementary attention map:

\begin{equation}
    \mathbf{z}^{q,l}
    =
    \operatorname{Pool}(\mathbf{F}^{q},\mathbf{A}^{q}),
    \qquad
    \mathbf{z}^{q,b}
    =
    \operatorname{Pool}(\mathbf{F}^{q},1-\mathbf{A}^{q}),
\end{equation}

where $\operatorname{Pool}(\cdot)$ denotes normalized weighted pooling. Both embeddings are $\ell_2$-normalized before computing relations. Thus, each image is represented by a lesion-focused embedding $\mathbf{z}^{q,l}$ and a contextual embedding $\mathbf{z}^{q,b}$.

PLCRD contains four relational objectives. First, inter-lesion relation alignment encourages the student to preserve the teacher's similarity structure among lesion embeddings. For a mini-batch, the lesion relation matrix is computed as

\begin{equation}
    \mathbf{R}^{q,l} =
    \mathbf{Z}^{q,l}(\mathbf{Z}^{q,l})^{\top},
    \qquad q \in \{T,S\},
\end{equation}

where $\mathbf{Z}^{q,l}$ contains the normalized lesion embeddings in the mini-batch. The student is trained to match the teacher relation matrix, allowing the student to inherit the teacher's mask-informed lesion geometry.

Second, lesion-context affinity matching transfers the relationship between lesion evidence and surrounding context. For each image, the lesion-context affinity is computed as the cosine similarity between $\mathbf{z}^{q,l}$ and $\mathbf{z}^{q,b}$. Third, separation regularization prevents the student from collapsing lesion and context embeddings into an identical representation. Fourth, class-aware relational learning encourages lesion embeddings from the same diagnostic class to become closer while pushing apart embeddings from different classes.
The complete PLCRD objective is

\begin{equation}
    \mathcal{L}_{\mathrm{PLCRD}}
    =
    \mathcal{L}_{\mathrm{LR}}
    +
    \alpha\mathcal{L}_{\mathrm{LC}}
    +
    \beta\mathcal{L}_{\mathrm{SEP}}
    +
    \gamma\mathcal{L}_{\mathrm{CAR}},
\end{equation}

where $\mathcal{L}_{\mathrm{LR}}$ is the inter-lesion relation alignment loss, $\mathcal{L}_{\mathrm{LC}}$ is the lesion-context affinity matching loss, $\mathcal{L}_{\mathrm{SEP}}$ is the separation regularizer, and $\mathcal{L}_{\mathrm{CAR}}$ is the class-aware relational loss. The coefficients $\alpha$, $\beta$, and $\gamma$ balance the individual PLCRD components.

This relational formulation is suitable for heterogeneous teacher-student learning because it does not require identical feature dimensions. The student is not forced to copy teacher features directly; instead, it learns to preserve the diagnostic organization induced by the mask-aware teacher.

\subsection{Overall Optimization Objective}
\label{subsec:overall_objective}

The final student objective combines supervised classification, logit distillation, attention distillation, and PLCRD:

\begin{equation}
    \mathcal{L}_{\mathrm{total}}
    =
    \mathcal{L}_{\mathrm{cls}}
    +
    \lambda_{\mathrm{KD}}\mathcal{L}_{\mathrm{KD}}
    +
    \lambda_{\mathrm{att}}\mathcal{L}_{\mathrm{att}}
    +
    \lambda_{\mathrm{PLCRD}}\mathcal{L}_{\mathrm{PLCRD}},
\end{equation}

where $\lambda_{\mathrm{KD}}$, $\lambda_{\mathrm{att}}$, and $\lambda_{\mathrm{PLCRD}}$ control the relative strengths of the distillation objectives. During student training, the teacher is frozen, and gradients are applied only to the student.

\subsection{Inference}
\label{subsec:inference}

After training, the teacher and segmentation masks are discarded. The deployed model consists only of the image-only student. Given a test image $\mathbf{x}$, the student predicts its diagnostic class directly from the image. Therefore, PLCRD benefits from lesion masks during training but does not require ground-truth masks or a segmentation network during inference. This makes the framework practical for deployment in settings where lesion annotations are unavailable, expensive, or unreliable.

\section{Experimental Setup}
\label{sec:experiments}
\begin{figure*}[t]
    \centering
    \begin{subfigure}[b]{0.23\textwidth}
        \centering
        \includegraphics[width=\textwidth]{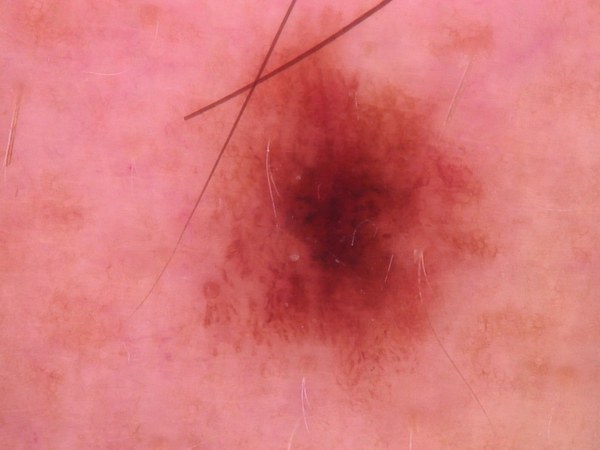}
        \caption{Sample 1}
    \end{subfigure}
    \hfill
    \begin{subfigure}[b]{0.23\textwidth}
        \centering
        \includegraphics[width=\textwidth]{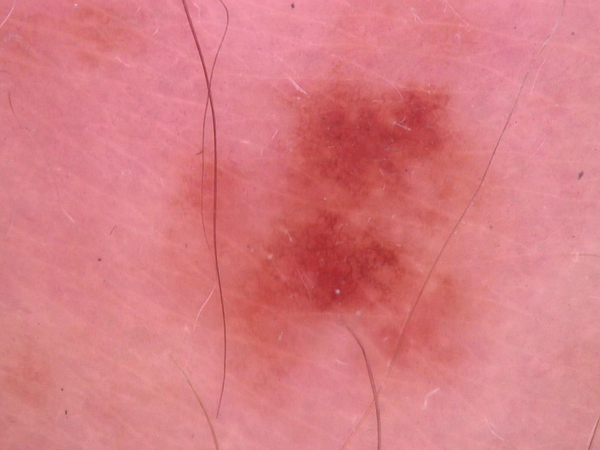}
        \caption{Sample 2}
    \end{subfigure}
    \hfill
    \begin{subfigure}[b]{0.23\textwidth}
        \centering
        \includegraphics[width=\textwidth]{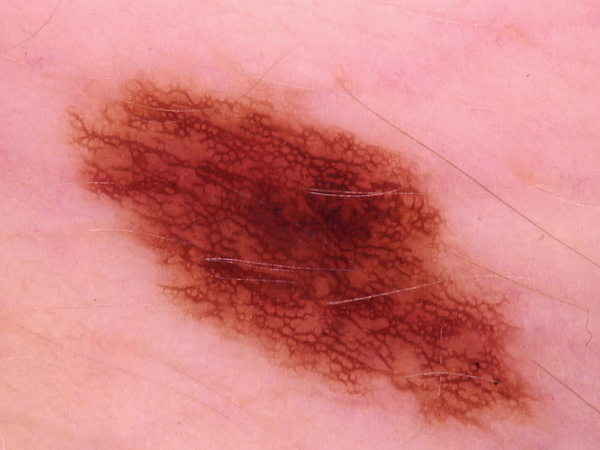}
        \caption{Sample 3}
    \end{subfigure}
    \hfill
    \begin{subfigure}[b]{0.23\textwidth}
        \centering
        \includegraphics[width=\textwidth]{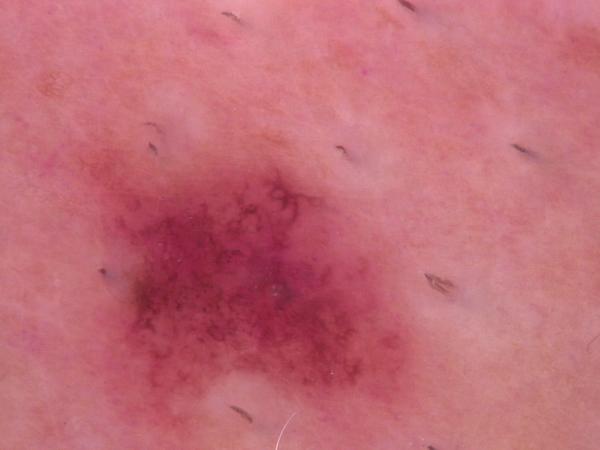}
        \caption{Sample 4}
    \end{subfigure}

    \vspace{0.6em}

    \begin{subfigure}[b]{0.23\textwidth}
        \centering
        \includegraphics[width=\textwidth]{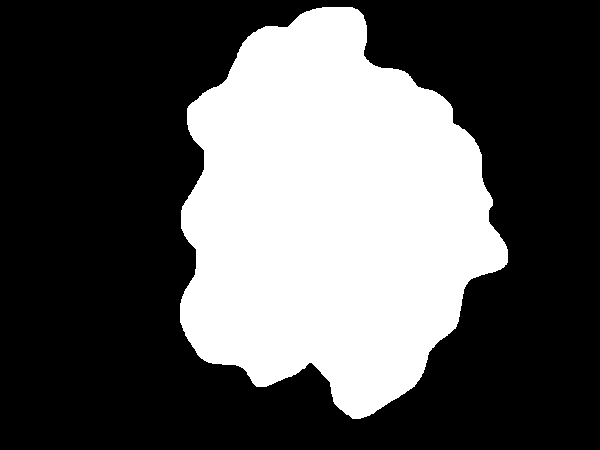}
        \caption{Segmentation 1}
    \end{subfigure}
    \hfill
    \begin{subfigure}[b]{0.23\textwidth}
        \centering
        \includegraphics[width=\textwidth]{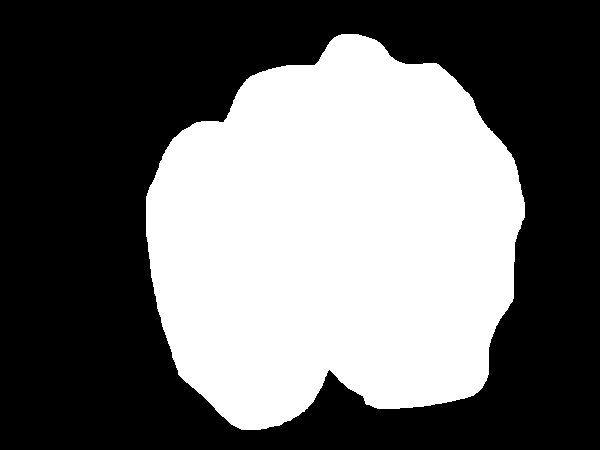}
        \caption{Segmentation 2}
    \end{subfigure}
    \hfill
    \begin{subfigure}[b]{0.23\textwidth}
        \centering
        \includegraphics[width=\textwidth]{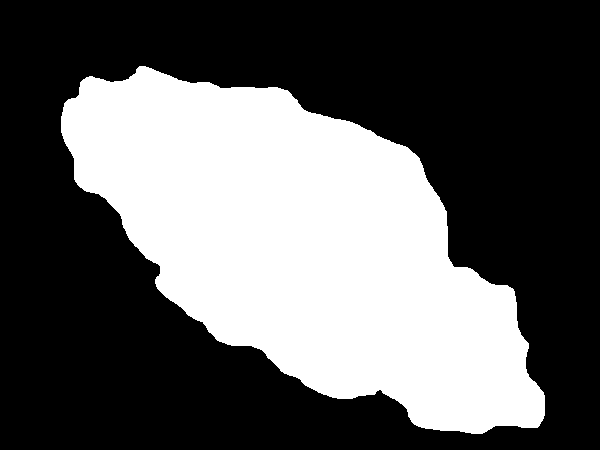}
        \caption{Segmentation 3}
    \end{subfigure}
    \hfill
    \begin{subfigure}[b]{0.23\textwidth}
        \centering
        \includegraphics[width=\textwidth]{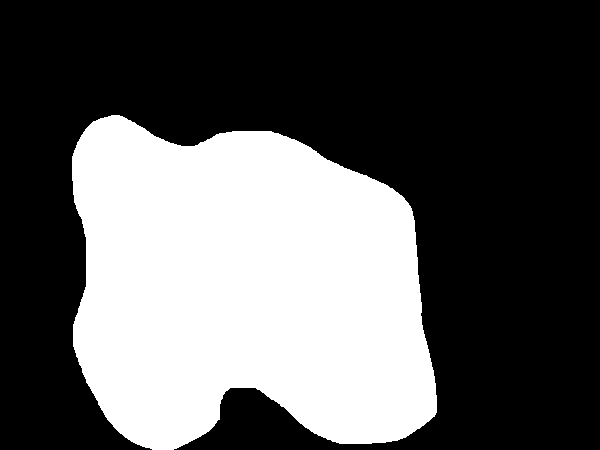}
        \caption{Segmentation 4}
    \end{subfigure}

    \caption{Representative dermoscopic samples from HAM10000 and their corresponding lesion segmentation masks. The first row shows four example dermoscopic images, and the second row shows the corresponding ground-truth lesion masks used as privileged supervision during training.}
    \label{fig:ham10000_samples_and_masks}
\end{figure*}
\subsection{Datasets}
\label{subsec:datasets}

Experiments were conducted using HAM10000 as the primary internal dataset and ISIC 2018 as the external validation dataset. HAM10000 contains 10,015 dermoscopic images from seven diagnostic categories and provides lesion-level metadata together with lesion segmentation masks~\cite{tschandl2018ham10000}. The segmentation masks were used only as privileged supervision during training and were not required during inference. The seven diagnostic categories are actinic keratoses and intraepithelial carcinoma (\texttt{akiec}), basal cell carcinoma (\texttt{bcc}), benign keratosis-like lesions (\texttt{bkl}), dermatofibroma (\texttt{df}), melanoma (\texttt{mel}), melanocytic nevi (\texttt{nv}), and vascular lesions (\texttt{vasc}). The class distribution is reported in Table~\ref{tab:class_distribution}.

External validation was performed on the ISIC 2018 test set without retraining or dataset-specific fine-tuning~\cite{codella2019skin}. This protocol evaluates whether the mask-informed knowledge learned from HAM10000 transfers to a different benchmark distribution. One locally unavailable image, \texttt{ISIC\_0035068}, was excluded from the external evaluation, resulting in 1,511 external test images. The dataset characteristics and internal data partitions are summarized in Table~\ref{tab:dataset_split_summary}.

\begin{table}[t]
\centering
\caption{Distribution of diagnostic classes in HAM10000.}
\label{tab:class_distribution}
\footnotesize
\setlength{\tabcolsep}{3pt}
\renewcommand{\arraystretch}{1.05}
\begin{tabular}{p{0.52\columnwidth}cc}
\toprule
Class & Abbreviation & Images \\
\midrule
Actinic keratoses & \texttt{akiec} & 327 \\
Basal cell carcinoma & \texttt{bcc} & 514 \\
Benign keratosis-like lesions & \texttt{bkl} & 1,099 \\
Dermatofibroma & \texttt{df} & 115 \\
Melanoma & \texttt{mel} & 1,113 \\
Melanocytic nevi & \texttt{nv} & 6,705 \\
Vascular lesions & \texttt{vasc} & 142 \\
\midrule
Total & -- & 10,015 \\
\bottomrule
\end{tabular}
\end{table}

\begin{table*}[t]
\centering
\caption{Dataset characteristics and lesion-disjoint split summary. HAM10000 was used for internal training, validation, and testing, while ISIC 2018 was used only for external evaluation without retraining.}
\label{tab:dataset_split_summary}
\begin{tabular}{lccccc}
\toprule
Dataset / Partition & Images & Lesions & Classes & Masks Available & Usage \\
\midrule
HAM10000 total & 10,015 & 7,470 & 7 & Yes & Internal dataset \\
Training split & 6,008 & 4,481 & 7 & Yes & Model training \\
Validation split & 2,003 & 1,494 & 7 & Yes & Model selection \\
Test split & 2,004 & 1,495 & 7 & Yes & Internal evaluation \\
ISIC 2018 external & 1,511 & -- & 7 & No & External evaluation \\
\bottomrule
\end{tabular}
\end{table*}
\begin{table}[t]
\centering
\caption{Preprocessing and augmentation settings.}
\label{tab:preprocessing}
\footnotesize
\setlength{\tabcolsep}{3pt}
\renewcommand{\arraystretch}{1.08}
\begin{tabular}{p{0.31\columnwidth}p{0.62\columnwidth}}
\toprule
Setting & Configuration \\
\midrule
Input &
RGB images resized to $224 \times 224$ pixels \\
Training crop &
Random resized crop with scale $0.75$--$1.00$ and
aspect ratio $0.85$--$1.15$ \\
Flipping &
Horizontal and vertical flips, each with $p=0.5$ \\
Rotation &
Random rotation up to $\pm180^\circ$ with reflect
padding and $p=0.5$ \\
Color jitter &
Brightness and contrast $=0.15$, saturation $=0.10$,
hue $=0.03$, and $p=0.5$ \\
Evaluation &
Deterministic resize to $224 \times 224$ pixels \\
Normalization &
Mean $(0.485,0.456,0.406)$ and standard deviation
$(0.229,0.224,0.225)$ \\
Mask processing &
Intensity scaled to $[0,1]$ with at least 64 lesion
pixels retained \\
Crop retry &
Up to 10 attempts, followed by deterministic-resize
fallback \\
\bottomrule
\end{tabular}
\end{table}

\begin{table}[t]
\centering
\caption{Training configuration.}
\label{tab:training_configuration}
\footnotesize
\setlength{\tabcolsep}{3pt}
\renewcommand{\arraystretch}{1.08}
\begin{tabular}{p{0.38\columnwidth}p{0.55\columnwidth}}
\hline
Setting & Configuration \\
\hline
Input size & $224 \times 224$ \\
Teacher input & Image and mask-guided lesion view \\
Student input & Image only \\
Loss for teacher & Classification and localization losses \\
Loss for student & Classification, logit KD, attention distillation, and PLCRD \\
Class imbalance & Class-balanced cross-entropy \\
Model selection & Best validation checkpoint \\
Internal evaluation & HAM10000 held-out test fold \\
External evaluation & ISIC 2018 without retraining \\
Inference model & Student only \\
\hline
\end{tabular}
\end{table}
\subsection{Data Partitioning and Leakage Prevention}
\label{subsec:data_partitioning}

To avoid data leakage, HAM10000 was partitioned at the lesion level rather than the image level. Images belonging to the same lesion were assigned to the same partition. This is important because HAM10000 contains multiple images for some lesions, and image-level random splitting could otherwise place visually similar images of the same lesion into both training and test sets. A five-fold lesion-disjoint partitioning protocol was used. In the reported setting, fold 0 was used for testing, fold 1 for validation, and folds 2--4 for training.
The validation set was used for model selection and early stopping, while the test set was used only for final internal evaluation. External validation on ISIC 2018 was performed after training on HAM10000, without using ISIC 2018 images for optimization, model selection, or hyperparameter tuning.

\subsection{Preprocessing and Data Augmentation}
\label{subsec:preprocessing}

All dermoscopic images were converted to RGB and resized to $224 \times 224$ pixels. During training, random resized cropping, horizontal and vertical flipping, rotation, and color jitter were applied to improve robustness to scale, orientation, and acquisition variability. During validation and testing, deterministic resizing was used. Image intensities were normalized using ImageNet mean and standard deviation values. Segmentation masks were resized consistently with the corresponding images and converted to binary maps. Masks with fewer than 64 lesion pixels were excluded from mask-based supervision to avoid unstable privileged guidance. The preprocessing and augmentation settings are summarized in Table~\ref{tab:preprocessing}.

\begin{table*}[t]
\centering
\caption{\small Internal and external lesion-level performance comparison. Internal results are computed on the held-out HAM10000 test split, and external results are computed on ISIC 2018 without retraining. Values are reported as mean $\pm$ standard deviation across the completed runs. The best result in each column is shown in bold, and the second-best result is underlined.}
\label{tab:internal_external_lesion_level_results}
\resizebox{\textwidth}{!}{
\begin{tabular}{lcccccccc}
\toprule \midrule
\multirow{2}{*}{Method}
& \multicolumn{4}{c}{Internal HAM10000 Lesion-Level Performance}
& \multicolumn{4}{c}{External ISIC 2018 Lesion-Level Performance} \\
\cmidrule(lr){2-5}
\cmidrule(lr){6-9}
& \makecell{Balanced\\Accuracy ($\uparrow$)}
& Macro-F1 ($\uparrow$)
& AUROC ($\uparrow$)
& Accuracy ($\uparrow$)
& \makecell{Balanced\\Accuracy ($\uparrow$)}
& Macro-F1 ($\uparrow$)
& AUROC ($\uparrow$)
& Accuracy ($\uparrow$) \\ 
\midrule
ResNet-50
& 0.700 $\pm$ 0.016
& 0.688 $\pm$ 0.024
& 0.963 $\pm$ 0.001
& 0.839 $\pm$ 0.011
& 0.689 $\pm$ 0.022
& 0.692 $\pm$ 0.007
& 0.952 $\pm$ 0.001
& 0.814 $\pm$ 0.011 \\ \midrule

DenseNet-121
& 0.709 $\pm$ 0.011
& 0.725 $\pm$ 0.016
& 0.972 $\pm$ 0.001
& 0.869 $\pm$ 0.005
& 0.694 $\pm$ 0.014
& 0.716 $\pm$ 0.018
& 0.960 $\pm$ 0.003
& 0.834 $\pm$ 0.009 \\ \midrule

EfficientNet-B0
& 0.739 $\pm$ 0.019
& 0.745 $\pm$ 0.015
& 0.969 $\pm$ 0.005
& 0.878 $\pm$ 0.005
& 0.686 $\pm$ 0.007
& 0.706 $\pm$ 0.004
& 0.948 $\pm$ 0.003
& 0.836 $\pm$ 0.005 \\ \midrule

EfficientNet-B3
& 0.703 $\pm$ 0.003
& 0.712 $\pm$ 0.024
& 0.964 $\pm$ 0.004
& 0.864 $\pm$ 0.016
& 0.686 $\pm$ 0.019
& 0.699 $\pm$ 0.017
& 0.947 $\pm$ 0.001
& 0.828 $\pm$ 0.009 \\ \midrule

ConvNeXt-Tiny
& \underline{0.743 $\pm$ 0.027}
& 0.739 $\pm$ 0.027
& {0.974 $\pm$ 0.003}
& \underline{0.881 $\pm$ 0.010}
& \underline{0.716 $\pm$ 0.029}
& 0.727 $\pm$ 0.011
& 0.962 $\pm$ 0.002
& \underline{0.838 $\pm$ 0.008} \\ \midrule

MobileNetV3-Small
& 0.698 $\pm$ 0.008
& 0.701 $\pm$ 0.014
& 0.964 $\pm$ 0.002
& 0.844 $\pm$ 0.008
& 0.627 $\pm$ 0.017
& 0.636 $\pm$ 0.008
& 0.950 $\pm$ 0.005
& 0.779 $\pm$ 0.014 \\ \midrule

EfficientFormerV2-S0
& 0.723 $\pm$ 0.016
& 0.714 $\pm$ 0.029
& 0.967 $\pm$ 0.004
& 0.858 $\pm$ 0.008
& 0.682 $\pm$ 0.012
& 0.682 $\pm$ 0.003
& 0.954 $\pm$ 0.007
& 0.816 $\pm$ 0.005 \\ \midrule

ViT-Small/16
& 0.698 $\pm$ 0.043
& 0.706 $\pm$ 0.013
& 0.966 $\pm$ 0.003
& 0.859 $\pm$ 0.015
& 0.676 $\pm$ 0.013
& 0.685 $\pm$ 0.019
& 0.956 $\pm$ 0.002
& 0.807 $\pm$ 0.029 \\ \midrule

DeiT-Small/16
& 0.683 $\pm$ 0.003
& 0.693 $\pm$ 0.007
& 0.962 $\pm$ 0.003
& 0.852 $\pm$ 0.004
& 0.653 $\pm$ 0.008
& 0.659 $\pm$ 0.001
& 0.952 $\pm$ 0.002
& 0.790 $\pm$ 0.010 \\ \midrule

Swin-Tiny
& 0.733 $\pm$ 0.020
& \underline{0.754 $\pm$ 0.006}
& \underline{0.975 $\pm$ 0.001}
& \textbf{0.888 $\pm$ 0.004}
& 0.695 $\pm$ 0.004
& \underline{0.730 $\pm$ 0.002}
& \underline{0.963 $\pm$ 0.002}
& \textbf{0.853 $\pm$ 0.001} \\ \midrule

\textbf{PLCRD (Proposed)}
& \textbf{0.764 $\pm$ 0.023}
& \textbf{0.773 $\pm$ 0.018}
& \textbf{0.976 $\pm$ 0.002}
& 0.878 $\pm$ 0.013
& \textbf{0.722 $\pm$ 0.020}
& \textbf{0.732 $\pm$ 0.008}
& \textbf{0.967 $\pm$ 0.002}
& 0.835 $\pm$ 0.010 \\ \midrule
\bottomrule
\end{tabular}
}
\end{table*}

\subsection{Compared Methods}
\label{subsec:compared_methods}

We compared PLCRD with a broad set of image-only, segmentation-dependent, localization-guided, and distillation-based baselines. These comparisons were designed to evaluate whether the proposed gains arise from privileged lesion-context relational transfer rather than from stronger backbones, direct mask usage, localization supervision, or conventional teacher-student learning. The first group includes standard image-only classifiers trained without mask information, representing the practical deployment setting in which only dermoscopic images are available. We also evaluated several representative backbone architectures, including convolutional, lightweight, and transformer-based classifiers, to assess whether PLCRD improves over strong image-only alternatives.

The second group contains segmentation-dependent variants. The ground-truth region-of-interest (GT-ROI) oracle uses lesion masks at inference and therefore provides a mask-dependent reference. The predicted-mask ROI baseline follows the common two-stage strategy in which a lesion segmentation model first estimates the lesion region and the classifier then operates on the resulting ROI~\cite{mahbod2020effects,yi2019integrated}. We further included a joint segmentation-classification baseline to evaluate whether shared feature learning across segmentation and diagnostic tasks improves classification performance~\cite{yang2017novel,song2020end}. These baselines test whether directly using lesion masks or segmentation branches is preferable to using masks only as privileged training information.

The third group consists of distillation and localization-guided baselines. Conventional knowledge distillation trains an image-only student using softened predictions from a teacher~\cite{hinton2015distilling}. Privileged logit distillation uses a mask-aware teacher but transfers only the teacher's output distribution. Direct localization supervision trains the student with an auxiliary mask-based localization constraint, while attention distillation transfers the teacher's spatial attention to the student~\cite{zagoruyko2017paying}. We also evaluated spatial relational distillation, which transfers generic relation-level information between teacher and student representations, following the broader idea of relational knowledge distillation~\cite{park2019relational}. Finally, the PMD baseline combines privileged teacher guidance with spatial evidence transfer but does not explicitly model the lesion-context relational structure proposed in PLCRD.

PLCRD differs from all these baselines in both training and deployment. During training, lesion masks are used only to construct a privileged mask-aware teacher. The final student learns from softened diagnostic predictions, class-conditioned attention, and lesion-context relational knowledge. During inference, PLCRD requires only the original dermoscopic image and does not require ground-truth masks, predicted masks, or an auxiliary segmentation network. This design allows the model to benefit from mask-informed supervision while preserving a fully image-only deployment pathway.

\subsection{Implementation Details}
\label{subsec:implementation_details}

The privileged teacher was trained using both the original dermoscopic image and the corresponding mask-guided lesion view, whereas the student was trained using only the original image. The teacher used a stronger backbone to provide mask-informed supervision, and the deployable student used a lightweight image-classification backbone to preserve inference efficiency. In our main configuration, the teacher was implemented with Swin-Tiny and the student with EfficientFormerV2-S0. The teacher was first optimized using class-balanced classification loss together with mask-aware localization supervision. After teacher training, the teacher parameters were frozen, and the student was optimized using supervised classification, logit distillation, class-conditioned attention distillation, and the proposed PLCRD objective.
All images were resized to $224 \times 224$ pixels, and all models were trained with class-balanced cross-entropy to reduce the effect of the highly imbalanced class distribution. The student distillation objective used a temperature of $\tau=4$. The loss weights were set to $\lambda_{\mathrm{KD}}=0.75$, $\lambda_{\mathrm{att}}=0.10$, and $\lambda_{\mathrm{PLCRD}}=0.10$. Within the PLCRD objective, the component weights were set to $\alpha=0.25$, $\beta=0.10$, and $\gamma=0.25$. Models were trained for up to 80 epochs using validation-based checkpoint selection. The best checkpoint was selected according to validation performance and then evaluated on the held-out HAM10000 test split and the ISIC 2018 external set.

All experiments were conducted using fixed random seeds for reproducibility. The internal results were reported on the lesion-disjoint HAM10000 test split, and external results were obtained on ISIC 2018 without retraining or dataset-specific fine-tuning. During inference, only the student backbone and its classification head were retained; the privileged teacher, lesion masks, attention-transfer objective, and PLCRD losses were used only during training.

\begin{table*}[t]
\centering
\caption{\small Component-wise ablation of mask usage, localization supervision, and
distillation strategies. Internal results are computed on the held-out
HAM10000 test split, while external results are computed on ISIC 2018 without
retraining. Values are reported as mean $\pm$ standard deviation across the
completed runs. A check mark indicates that the corresponding component is
enabled. The best result in each column is shown in bold, and the second-best
result is underlined.}
\label{tab:method_ablation}

\setlength{\tabcolsep}{3.2pt}
\renewcommand{\arraystretch}{1.15}

\resizebox{\textwidth}{!}{%
\begin{tabular}{lcccccccccccc}
\toprule \midrule
\multirow{2}{*}{Method}
& \multicolumn{8}{c}{Training Components}
& \multicolumn{2}{c}{Internal HAM10000}
& \multicolumn{2}{c}{External ISIC 2018} \\
\cmidrule(lr){2-9}
\cmidrule(lr){10-11}
\cmidrule(lr){12-13}

& \makecell{ROI\\Input}
& \makecell{Joint\\Seg.}
& \makecell{Conv.\\KD}
& \makecell{Priv.\\KD}
& \makecell{Spatial\\Relation}
& \makecell{Attention\\Distill.}
& \makecell{Direct\\Loc.}
& \makecell{PLCRD\\Relation}
& \makecell{Balanced\\Accuracy}
& \makecell{Macro-\\F1}
& \makecell{Balanced\\Accuracy}
& \makecell{Macro-\\F1} \\
\midrule

\textbf{PLCRD (Proposed)}
& -- & -- & -- & $\checkmark$ & $\checkmark$
& $\checkmark$ & -- & $\checkmark$
& \textbf{0.764 $\pm$ 0.023}
& \textbf{0.773 $\pm$ 0.018}
& \textbf{0.722 $\pm$ 0.020}
& \textbf{0.732 $\pm$ 0.008} \\ \midrule

\quad Image-only student
& -- & -- & -- & -- & -- & -- & -- & --
& \underline{0.720 $\pm$ 0.002}
& 0.704 $\pm$ 0.002
& \underline{0.683 $\pm$ 0.018}
& \underline{0.683 $\pm$ 0.016} \\ \midrule

\quad Ground-truth ROI
& $\checkmark$ & -- & -- & -- & -- & -- & -- & --
& 0.657 $\pm$ 0.016
& 0.669 $\pm$ 0.014
& --
& -- \\ \midrule

\quad Predicted-mask ROI
& $\checkmark$ & -- & -- & -- & -- & -- & -- & --
& 0.638 $\pm$ 0.021
& 0.666 $\pm$ 0.037
& 0.580 $\pm$ 0.013
& 0.620 $\pm$ 0.008 \\ \midrule

\quad Joint seg-classification
& -- & $\checkmark$ & -- & -- & -- & -- & -- & --
& \underline{0.720 $\pm$ 0.037}
& 0.717 $\pm$ 0.061
& 0.625 $\pm$ 0.027
& 0.624 $\pm$ 0.052 \\ \midrule
 
\quad Conventional KD
& -- & -- & $\checkmark$ & -- & -- & -- & -- & --
& 0.700 $\pm$ 0.005
& 0.714 $\pm$ 0.001
& 0.643 $\pm$ 0.042
& 0.671 $\pm$ 0.025 \\ \midrule

\quad Direct LS
& -- & -- & -- & -- & -- & -- & $\checkmark$ & --
& 0.701 $\pm$ 0.031
& 0.691 $\pm$ 0.015
& 0.625 $\pm$ 0.012
& 0.619 $\pm$ 0.008 \\ \midrule

\quad Privileged LD
& -- & -- & -- & $\checkmark$ & -- & -- & -- & --
& 0.696 $\pm$ 0.003
& 0.727 $\pm$ 0.007
& 0.629 $\pm$ 0.037
& 0.663 $\pm$ 0.026 \\ \midrule

\quad Spatial RD
& -- & -- & -- & -- & $\checkmark$ & -- & -- & --
& 0.700 $\pm$ 0.019
& 0.703 $\pm$ 0.020
& 0.659 $\pm$ 0.002
& 0.674 $\pm$ 0.027 \\ \midrule

\quad SL variant
& -- & -- & -- & $\checkmark$ & $\checkmark$ & --
& $\checkmark$ & --
& 0.675 $\pm$ 0.022
& 0.707 $\pm$ 0.015
& 0.639 $\pm$ 0.019
& 0.676 $\pm$ 0.016 \\ \midrule

\quad Attention distillation
& -- & -- & -- & -- & -- & $\checkmark$ & -- & --
& 0.699 $\pm$ 0.019
& 0.676 $\pm$ 0.030
& 0.643 $\pm$ 0.019
& 0.643 $\pm$ 0.018 \\ \midrule

\quad Localization-guided KD
& -- & -- & -- & $\checkmark$ & -- & --
& $\checkmark$ & --
& \underline{0.720 $\pm$ 0.023}
& \underline{0.745 $\pm$ 0.017}
& 0.639 $\pm$ 0.034
& 0.673 $\pm$ 0.022 \\ \midrule

\quad PMD baseline
& -- & -- & -- & $\checkmark$ & $\checkmark$
& $\checkmark$ & -- & --
& 0.704 $\pm$ 0.028
& 0.722 $\pm$ 0.030
& 0.641 $\pm$ 0.003
& 0.664 $\pm$ 0.003 \\ \midrule

\quad PLCRD relation
& -- & -- & -- & -- & -- & -- & -- & $\checkmark$
& 0.710 $\pm$ 0.015
& 0.679 $\pm$ 0.024
& 0.676 $\pm$ 0.032
& 0.661 $\pm$ 0.002 \\ \midrule
\bottomrule
\end{tabular}%
}
\end{table*}

\subsection{Evaluation Metrics}
\label{subsec:evaluation_metrics}

Classification performance was evaluated using accuracy, balanced accuracy, macro-averaged F1-score, macro-averaged area under the receiver operating characteristic curve, sensitivity, specificity, and Matthews correlation coefficient. Because HAM10000 is class-imbalanced, balanced accuracy and macro-F1 were treated as primary performance indicators. Macro-averaged metrics give equal importance to all classes and are therefore more informative than overall accuracy alone.
In addition to image-level evaluation, lesion-level evaluation was performed by aggregating predictions from images belonging to the same lesion. For each lesion, predicted class probabilities were averaged across its associated images, and the final lesion label was obtained from the highest averaged probability. Calibration was assessed using negative log-likelihood, Brier score, and expected calibration error. Localization behavior was evaluated using attention-based measures, including lesion attention concentration and pointing-game accuracy. Robustness was assessed by evaluating model performance under common image corruptions and acquisition perturbations.

\subsection{Statistical Analysis}
\label{subsec:statistical_analysis}

Statistical analysis was conducted to determine whether the proposed PLCRD framework provided consistent improvements over the main comparison methods. Paired comparisons were performed using test predictions from the same data partitions. Confidence intervals were estimated using bootstrap resampling. For metric-level comparisons, paired non-parametric tests were used when normality assumptions were not appropriate. Multiple-comparison correction was applied when several related hypotheses were tested. The main statistical comparisons focused on PLCRD against the image-only classifier and the strongest mask-free distillation baseline.

\section{Results and Discussion}
\label{sec:results}

\subsection{Internal and External Lesion-Level Classification}
\label{subsec:classification_results}

Table~\ref{tab:internal_external_lesion_level_results} compares the proposed PLCRD framework with representative convolutional, lightweight, and transformer-based classifiers on the held-out HAM10000 test split and the external ISIC 2018 dataset. On HAM10000, PLCRD achieves the highest balanced accuracy of $0.764 \pm 0.023$, Macro-F1 of $0.773 \pm 0.018$, and AUROC of $0.976 \pm 0.002$. These results outperform the strongest conventional baselines in the class-balanced metrics, including ConvNeXt-Tiny in balanced accuracy and Swin-Tiny in Macro-F1 and AUROC. Although Swin-Tiny obtains the highest overall accuracy of $0.888 \pm 0.004$, PLCRD achieves a more favorable balance across diagnostic categories, which is particularly important for the highly imbalanced HAM10000 dataset. The improvement in balanced accuracy and Macro-F1 suggests that the proposed lesion-context relational supervision helps the model reduce majority-class bias and better recognize underrepresented lesion categories.
The external results further confirm the generalization capability of PLCRD. Without any retraining on ISIC 2018, the proposed model achieves the best balanced accuracy of $0.722 \pm 0.020$, Macro-F1 of $0.732 \pm 0.008$, and AUROC of $0.967 \pm 0.002$. PLCRD maintains competitive overall accuracy while outperforming the compared backbones in the metrics that place equal importance on each diagnostic class. The decrease from the internal to the external dataset is also relatively controlled, with balanced accuracy declining by $0.042$ and Macro-F1 by $0.041$. This indicates that the model retains most of its class-balanced discrimination under cross-dataset variation. Overall, the internal and external evaluations show that PLCRD provides a strong combination of diagnostic performance, minority-class sensitivity, and robustness to changes in image characteristics and data distribution.

\subsection{Method-Level Comparison and Ablation Analysis}
\label{subsec:method_ablation}

Table~\ref{tab:method_ablation} compares PLCRD with alternative strategies for incorporating lesion masks, localization cues, and privileged supervision. Direct use of ground-truth or predicted-mask ROI inputs does not improve classification, with the predicted-mask ROI producing the weakest internal balanced accuracy of $0.638 \pm 0.021$ and external balanced accuracy of $0.580 \pm 0.013$. Similarly, joint segmentation-classification and direct localization supervision provide only moderate gains, indicating that explicitly forcing the classifier to reproduce lesion masks is not sufficient for robust diagnosis. Conventional KD, privileged logit distillation, spatial relation distillation, and attention distillation offer complementary improvements, but none consistently dominates across both datasets. Among the non-PLCRD alternatives, localization-guided KD achieves the strongest internal Macro-F1 of $0.745 \pm 0.017$, while the image-only student provides the strongest external performance with balanced accuracy and Macro-F1 of $0.683 \pm 0.018$ and $0.683 \pm 0.016$, respectively.

The full PLCRD framework achieves the best performance in every reported ablation metric, reaching internal balanced accuracy and Macro-F1 values of $0.764 \pm 0.023$ and $0.773 \pm 0.018$, together with external values of $0.722 \pm 0.020$ and $0.732 \pm 0.008$. In comparison, the PMD baseline achieves an internal Macro-F1 of $0.722 \pm 0.030$ and an external Macro-F1 of $0.664 \pm 0.003$, while the isolated PLCRD relation variant reaches $0.679 \pm 0.024$ and $0.661 \pm 0.002$, respectively. These results show that privileged distillation and lesion-context relational learning are most effective when combined. The complete formulation improves both internal recognition and external generalization, suggesting that PLCRD transfers not only lesion-focused supervision but also diagnostically meaningful relations between the lesion and its surrounding context.

\subsection{Magnitude of Performance Improvements}
\label{subsec:effect_magnitude}
\begin{table}[t]
\centering
\caption{\small Observed internal Macro-F1 improvements of PLCRD over competing
methods on the held-out HAM10000 test split. The absolute difference is
defined as the Macro-F1 of PLCRD minus that of the comparison method.
Relative improvement is calculated with respect to the comparison method.
These results represent descriptive effect estimates rather than statistical
significance tests.}
\label{tab:observed_effects}
\resizebox{\columnwidth}{!}{
\setlength{\tabcolsep}{7.0pt}
\renewcommand{\arraystretch}{1.0}
\begin{tabular}{lcccc}
\toprule
Comparison with PLCRD
& \makecell{Macro-F1}
& \makecell{PLCRD\\Macro-F1}
& \makecell{Absolute\\Difference}
& \makecell{Relative\\Improvement} \\
\midrule
 vs.\ image-only
& 0.704
& 0.773
& $+0.069$
& $+9.80\%$ \\ \midrule
 vs.\ conventional KD
& 0.714
& 0.773
& $+0.059$
& $+8.26\%$ \\ \midrule
 vs.\ PMD 
& 0.722
& 0.773
& $+0.051$
& $+7.06\%$ \\ \midrule
 vs.\ JSC
& 0.717
& 0.773
& $+0.056$
& $+7.81\%$ \\ \midrule
 vs.\ DLS
& 0.691
& 0.773
& \textbf{$+0.082$}
& \textbf{$+11.87\%$} \\

\bottomrule
\end{tabular}}
\end{table}
Table~\ref{tab:observed_effects} summarizes the absolute and relative Macro-F1 improvements achieved by PLCRD over the principal competing strategies. PLCRD improves Macro-F1 by $0.069$ over the image-only student, $0.059$ over conventional KD, $0.051$ over PMD, and $0.056$ over joint segmentation-classification, corresponding to relative improvements between $7.06\%$ and $9.80\%$. The largest gain is observed over direct localization supervision, where PLCRD improves Macro-F1 by $0.082$, equivalent to an $11.87\%$ relative increase. These improvements indicate that the proposed framework provides benefits beyond isolated mask guidance, localization supervision, or conventional distillation by jointly transferring privileged lesion information and lesion-context relational structure.

\subsection{Localization and Quantitative Interpretability}
\label{subsec:localization_analysis}

Table~\ref{tab:localization_attention_quality} quantitatively evaluates how
the compared approaches distribute diagnostic attention relative to the
annotated lesion regions. Soft Dice measures the spatial agreement between
the continuous attention response and the complete lesion mask, while
pointing accuracy evaluates whether the maximum-response location lies inside
the lesion. Lesion-attention fraction measures the proportion of total
attention assigned to the lesion, whereas outside-attention fraction measures
attention leakage into the surrounding background. These metrics therefore
capture complementary aspects of localization quality rather than treating
the attention map as a conventional segmentation output.
PLCRD achieves the highest pointing accuracy of $0.964$ and lesion-attention
fraction of $0.956$, while reducing outside-lesion attention to $0.044$. Its
Soft Dice of $0.342$ is lower than the ROI-based alternatives because PLCRD
concentrates its responses on compact diagnostically informative regions
instead of reconstructing the complete lesion boundary. In contrast, the
ground-truth ROI and mask-ROI methods obtain stronger full-mask overlap but
require explicit or predicted lesion masks. Overall, the results indicate
that PLCRD learns highly lesion-centered diagnostic evidence while preserving
mask-free inference.

\begin{table}[t]
\centering
\caption{\small Localization and attention quality on the held-out HAM10000
test split. The best result in each column is shown in bold, and the
second-best result is underlined.}
\label{tab:localization_attention_quality}

\setlength{\tabcolsep}{5.0pt}
\renewcommand{\arraystretch}{1.15}

\resizebox{\columnwidth}{!}{%
\begin{tabular}{lcccc}
\toprule
Method
& \makecell{Soft Dice\\($\uparrow$)}
& \makecell{Pointing\\Accuracy ($\uparrow$)}
& \makecell{Lesion-Attention\\Fraction ($\uparrow$)}
& \makecell{Outside-Attention\\Fraction ($\downarrow$)} \\
\midrule

\textbf{PLCRD}
& 0.342
& \textbf{0.964}
& \textbf{0.956}
& \textbf{0.044} \\ \midrule

\quad Image-only
& 0.499
& 0.885
& 0.720
& 0.280 \\ \midrule

\quad Ground-truth ROI
& \textbf{0.610}
& 0.831
& \underline{0.734}
& \underline{0.266} \\ \midrule

\quad Mask ROI
& \underline{0.593}
& 0.875
& 0.733
& 0.267 \\ \midrule

\quad JSC
& 0.491
& \underline{0.904}
& 0.722
& 0.278 \\
\bottomrule
\end{tabular}%
}
\end{table}

\begin{figure*}[t]
    \centering

    \begin{subfigure}{\textwidth}
        \centering
        \includegraphics[width=\textwidth]{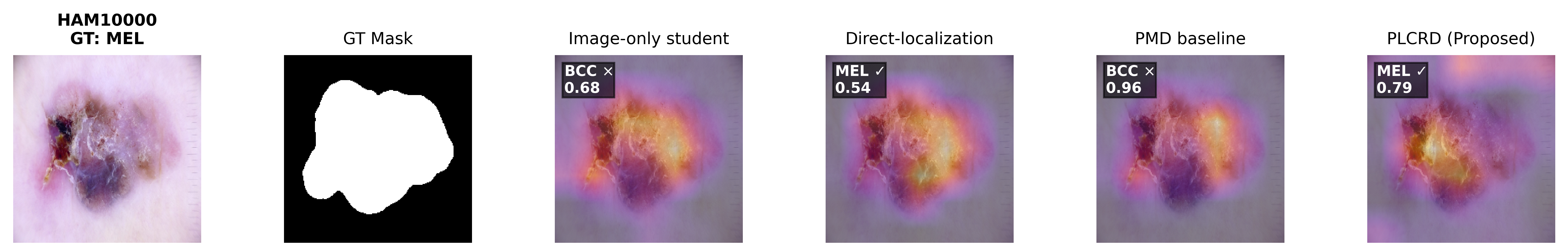}
        \caption{HAM10000 melanoma case. The image-only student and PMD
        misclassify the lesion as basal cell carcinoma, whereas direct
        localization and PLCRD correctly predict melanoma.}
        \label{fig:attention_mel}
    \end{subfigure}

    \vspace{4pt}

    \begin{subfigure}{\textwidth}
        \centering
        \includegraphics[width=\textwidth]{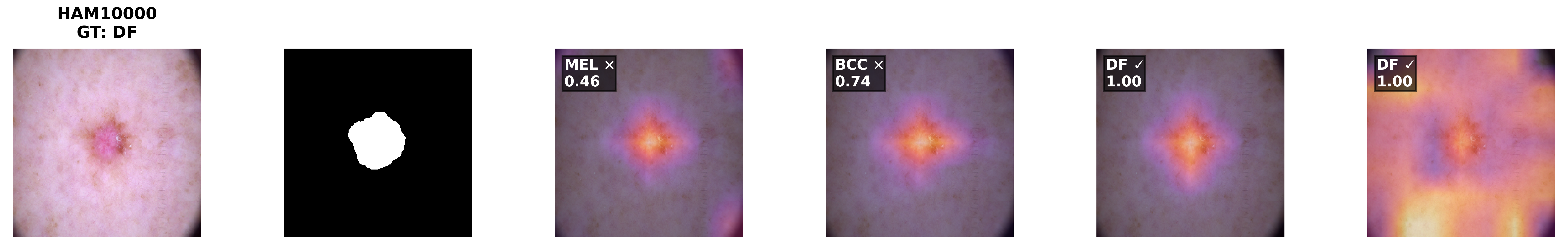}
        \caption{External ISIC 2018 melanocytic-nevus case evaluated without
        retraining. All methods correctly classify the lesion, while their
        attention maps exhibit different spatial distributions.}
        \label{fig:attention_external}
    \end{subfigure}

    \vspace{4pt}

    \begin{subfigure}{\textwidth}
        \centering
        \includegraphics[width=\textwidth]{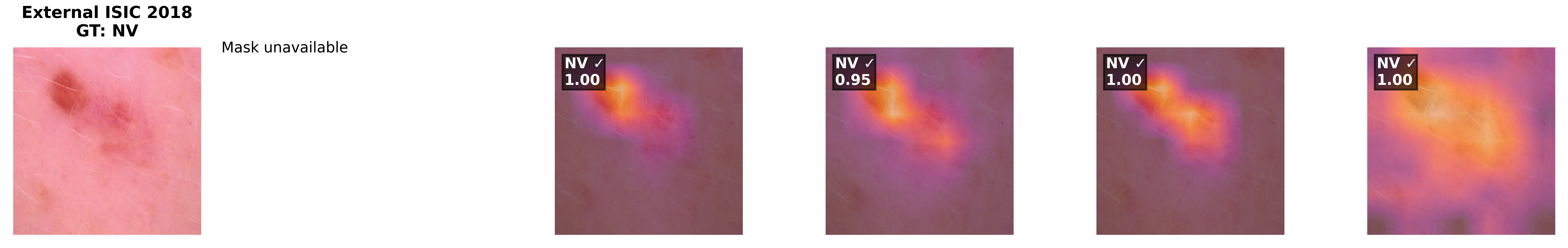}
        \caption{HAM10000 dermatofibroma case. The image-only and
        direct-localization models produce incorrect predictions, whereas
        PMD and PLCRD correctly recognize the rare lesion category.}
        \label{fig:attention_df}
    \end{subfigure}

    \caption{Qualitative comparison of class-discriminative Grad-CAM++
    visualizations for representative internal and external skin-lesion
    cases. Each row presents the input image, ground-truth mask when
    available, and the predictions, confidence scores, and attention maps
    produced by the image-only student, direct-localization model, PMD
    baseline, and PLCRD. The examples illustrate how privileged supervision
    and lesion-context relational learning redirect diagnostic evidence
    toward relevant lesion regions. The external ISIC 2018 example is
    evaluated without retraining, and PLCRD does not require a lesion mask
    during inference.}
    \label{fig:lesion_attention_comparison}
\end{figure*}

\subsection{Qualitative Lesion-Attention Analysis}
\label{subsec:qualitative_attention}

Figure~\ref{fig:lesion_attention_comparison} presents representative
Grad-CAM++ visualizations from the internal HAM10000 and external ISIC 2018
evaluations. In the melanoma example, the image-only student and PMD classify
the lesion as basal cell carcinoma, whereas direct localization and PLCRD
recover the correct melanoma diagnosis. The external ISIC 2018 example shows
that all methods correctly recognize the melanocytic nevus without
retraining, although the spatial extent of the attended regions differs
across models. In the dermatofibroma example, the image-only and
direct-localization models fail, while PMD and PLCRD correctly identify the
rare diagnostic category.
The qualitative maps further show that PLCRD generally concentrates evidence
within diagnostically relevant lesion regions while retaining limited
surrounding context. Unlike ROI-based approaches, the proposed method does
not require a lesion mask or segmentation model during deployment. The
visualizations therefore support the quantitative finding that PLCRD learns
lesion-centered classification evidence rather than reproducing a complete
segmentation mask.

\begin{figure*}[t]
    \centering
    \includegraphics[width=\textwidth]
    {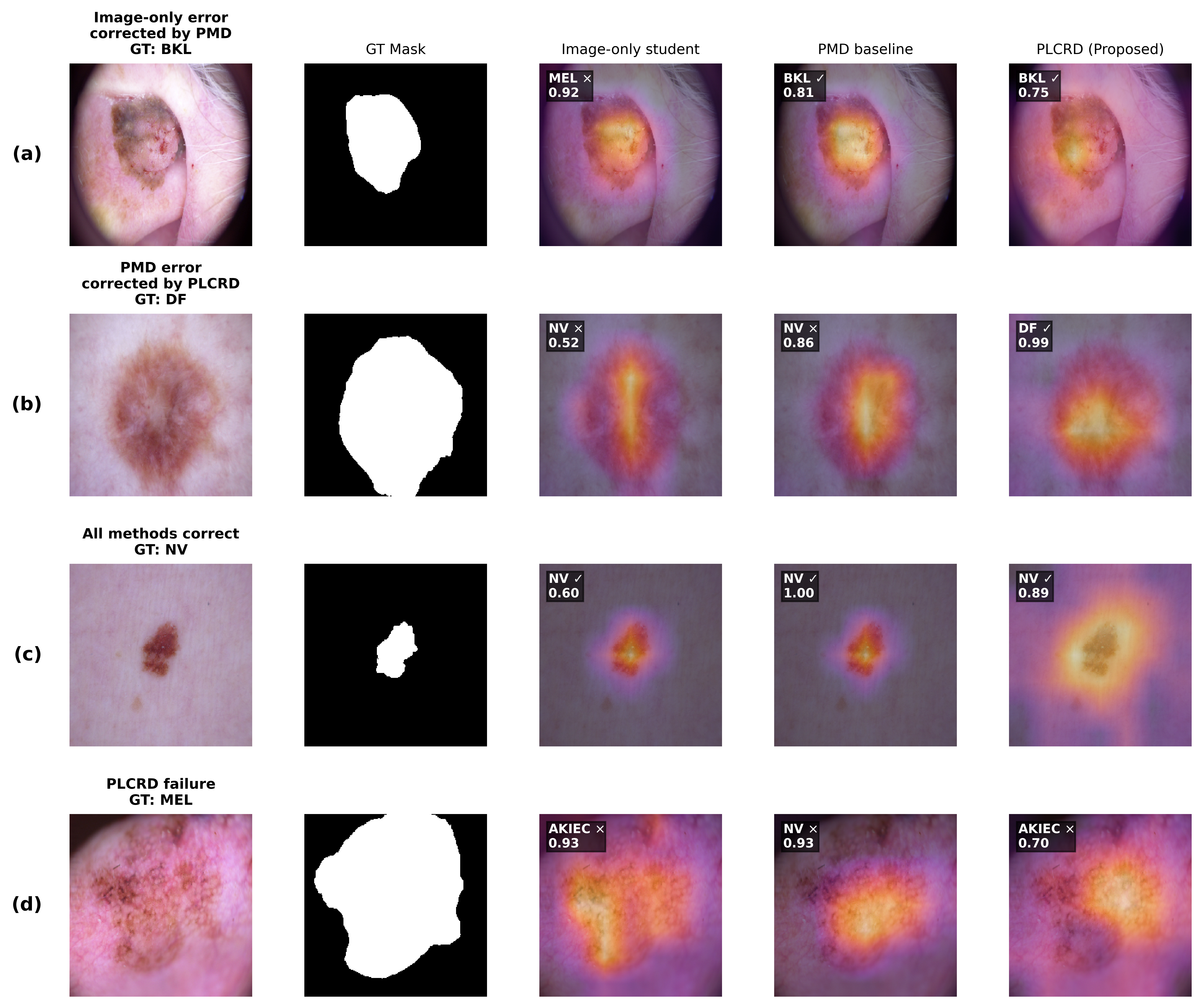}

    \caption{Case-based comparison of the image-only student, PMD baseline,
    and PLCRD. Panel (a) shows an image-only error corrected by both PMD and
    PLCRD. Panel (b) presents a PMD error corrected by PLCRD for a
    dermatofibroma case. Panel (c) shows a melanocytic-nevus example correctly
    classified by all methods. Panel (d) presents a remaining PLCRD failure,
    in which melanoma is confused with actinic keratosis. Each model panel
    reports the predicted class, confidence score, and class-discriminative
    Grad-CAM++ visualization.}
    \label{fig:corrected_predictions_failures}
\end{figure*}

\subsection{Case-Based Error Correction and Failure Analysis}
\label{subsec:case_analysis}

Figure~\ref{fig:corrected_predictions_failures} provides a case-based analysis
of corrected predictions and remaining errors. In Fig.~\ref{fig:corrected_predictions_failures}(a),
the image-only student misclassifies a benign keratosis-like lesion as
melanoma, whereas PMD and PLCRD recover the correct diagnosis. In
Fig.~\ref{fig:corrected_predictions_failures}(b), both the image-only student
and PMD classify a dermatofibroma as a nevus, while PLCRD correctly predicts
dermatofibroma with high confidence. These cases illustrate the complementary
benefit of privileged lesion supervision and lesion-context relational
learning.
Figure~\ref{fig:corrected_predictions_failures}(c) shows a melanocytic-nevus
case correctly classified by all methods, demonstrating that PLCRD preserves
already correct predictions. Figure~\ref{fig:corrected_predictions_failures}(d)
presents a remaining failure in which melanoma is confused with actinic
keratosis. This error highlights the difficulty of distinguishing lesions
with overlapping color, texture, and structural characteristics. Including
both corrected cases and an unresolved failure provides a balanced assessment
of the proposed method.

\subsection{Lesion-Context Relational Analysis}
\label{subsec:relation_analysis}

\begin{figure*}[t]
    \centering
    \includegraphics[width=\textwidth]
    {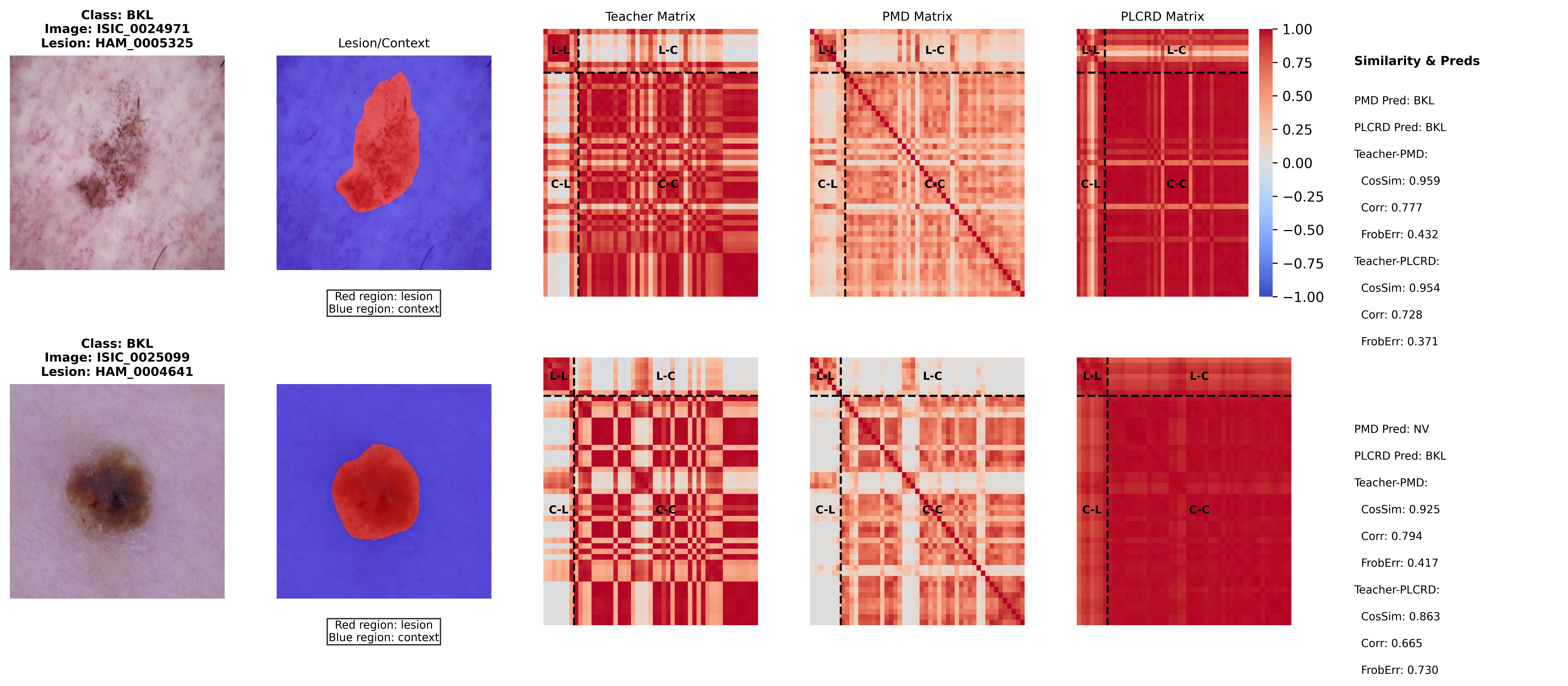}

    \caption{Visualization of lesion-context relational structure for two
    representative benign keratosis-like lesion cases. For each case, the
    input image and lesion-context partition are followed by the privileged
    teacher, PMD, and PLCRD relation matrices. The matrices are divided into
    lesion-to-lesion (L--L), lesion-to-context (L--C), context-to-lesion
    (C--L), and context-to-context (C--C) blocks. Cosine similarity,
    correlation, and normalized Frobenius error quantify agreement with the
    privileged teacher. The first case illustrates broadly consistent
    relational behavior across the compared methods, whereas the second
    provides a contrasting example in which PLCRD corrects the final
    prediction despite weaker global matrix similarity.}
    \label{fig:lesion_context_relations}
\end{figure*}

Figure~\ref{fig:lesion_context_relations} visualizes the relational structures
learned by the privileged teacher, PMD, and PLCRD. Each matrix is divided into
lesion-to-lesion, lesion-to-context, context-to-lesion, and
context-to-context blocks. These blocks describe how spatial representations
within the lesion interact with one another and with the surrounding skin
context. Cosine similarity, correlation, and normalized Frobenius error are
used to quantify agreement between the student relation matrices and the
privileged teacher.
The first example shows broadly consistent relational behavior across PMD and
PLCRD, with both methods producing the correct prediction. In the second
example, PMD predicts a melanocytic nevus, whereas PLCRD correctly predicts a
benign keratosis-like lesion despite weaker global matrix similarity under
some measures. This observation suggests that closer reconstruction of the
entire teacher relation matrix does not always correspond directly to better
classification. Instead, PLCRD may preserve task-relevant relational
patterns that are particularly useful for the final diagnostic decision.

\subsection{Calibration and Predictive Reliability}
\label{subsec:calibration}

Table~\ref{tab:calibration_reliability} evaluates whether the improvements in
classification performance are accompanied by reliable confidence estimates.
PLCRD achieves the lowest ECE of $0.022$, NLL of $0.358$, and Brier score of
$0.178$. These values are substantially lower than those of the image-only,
ROI-based, joint segmentation-classification, and ResNet-50 alternatives,
indicating improved agreement between predicted probabilities and observed
outcomes.
PLCRD simultaneously obtains the highest AUROC of $0.976$ and Macro-F1 of
$0.773$. Therefore, the improvement in predictive confidence is not obtained
at the expense of discrimination or class-balanced performance. The combined
results suggest that PLCRD produces both more accurate and better-calibrated
predictions, which is particularly important for clinical decision-support
settings where confidence estimates may influence referral or review
decisions.

\begin{table}[t]
\centering
\caption{\small Calibration and predictive reliability on the held-out
HAM10000 test split. The best result in each column is shown in bold, and the
second-best result is underlined.}
\label{tab:calibration_reliability}

\setlength{\tabcolsep}{6.0pt}
\renewcommand{\arraystretch}{1.15}

\resizebox{\columnwidth}{!}{%
\begin{tabular}{lccccc}
\toprule
Method
& ECE ($\downarrow$)
& NLL ($\downarrow$)
& Brier Score ($\downarrow$)
& AUROC ($\uparrow$)
& Macro-F1 ($\uparrow$) \\
\midrule

\textbf{PLCRD}
& \textbf{0.022}
& \textbf{0.358}
& \textbf{0.178}
& \textbf{0.976}
& \textbf{0.773} \\ \midrule

\quad Image-only
& \underline{0.030}
& \underline{0.652}
& \underline{0.344}
& \underline{0.949}
& 0.623 \\ \midrule

\quad Ground-truth ROI
& 0.032
& 0.755
& 0.397
& 0.938
& 0.552 \\ \midrule

\quad Mask ROI
& 0.052
& 0.724
& 0.380
& 0.931
& 0.568 \\ \midrule

\quad JSC
& 0.042
& 0.656
& 0.347
& 0.944
& \underline{0.603} \\ \midrule

\quad ResNet-50
& 0.056
& 0.788
& 0.396
& 0.921
& 0.543 \\
\bottomrule
\end{tabular}%
}
\end{table}

\subsection{Computational Complexity and Deployment Efficiency}
\label{subsec:efficiency}

Table~\ref{tab:model_efficiency} compares the computational requirements of
PLCRD with representative image-only, ROI-based, and joint
segmentation-classification approaches. PLCRD uses privileged lesion
information only during training, while the teacher, mask-related branches,
and distillation objectives are removed during deployment. Consequently, the
deployed model operates using an image-only inference path and does not
require an external mask or segmentation network.

PLCRD contains $0.931$ million parameters, requires $0.110$ GFLOPs, and has a
model size of $3.552$ MB. Its median CPU latency is $5.468$ ms per image, with
a $p_{95}$ latency of $7.158$ ms. These values are substantially lower than
those of ResNet-50, the mask-ROI pipeline, and the joint
segmentation-classification model. The results demonstrate that the proposed
training strategy introduces privileged supervision without increasing the
cost of the deployed classifier.

\begin{table}[t]
\centering
\caption{\small Model complexity and inference efficiency at an input
resolution of $224 \times 224$. FLOPs are measured for one forward pass. CPU
inference latency is measured after 100 warm-up iterations over 1,000 timed
iterations. PMD and PLCRD use the same lightweight image-only student during
deployment because the privileged teacher and distillation branches are
discarded after training.}
\label{tab:model_efficiency}

\setlength{\tabcolsep}{4.8pt}
\renewcommand{\arraystretch}{1.15}

\resizebox{\columnwidth}{!}{%
\begin{tabular}{lccccc}
\toprule
Method
& \makecell{Parameters\\(M)}
& \makecell{FLOPs\\(G)}
& \makecell{Model Size\\(MB)}
& \makecell{Median CPU\\Latency (ms)}
& \makecell{CPU $p_{95}$\\Latency (ms)} \\
\midrule

\textbf{PLCRD}
& \textbf{0.931}
& \textbf{0.110}
& \textbf{3.552}
& \textbf{5.468}
& \textbf{7.158} \\ \midrule

\quad Image-only
& \textbf{0.931}
& \textbf{0.110}
& \textbf{3.552}
& \textbf{5.468}
& \textbf{7.158} \\ \midrule

\quad ResNet-50
& 23.522
& 8.174
& 89.731
& 37.117
& 40.256 \\ \midrule

\quad Mask ROI
& 1.939
& 1.185
& 7.397
& 15.573
& 17.887 \\ \midrule

\quad JSC
& 1.272
& 8.666
& 4.853
& 39.560
& 42.917 \\ \midrule

\quad PMD
& \textbf{0.931}
& \textbf{0.110}
& \textbf{3.552}
& \textbf{5.468}
& \textbf{7.158} \\
\bottomrule
\end{tabular}%
}
\end{table}

\subsection{Overall Findings}
\label{subsec:overall_findings}

The experimental results consistently demonstrate that PLCRD improves
class-balanced skin-lesion classification while maintaining strong external
generalization. The method-level ablations show that direct ROI use,
localization supervision, conventional knowledge distillation, and isolated
relation learning do not individually reproduce the performance of the full
framework. The combination of privileged lesion supervision and
lesion-context relational distillation is therefore central to the observed
improvements.
The interpretability analysis further shows that PLCRD produces highly
lesion-centered diagnostic evidence, although it does not attempt to
reconstruct the complete lesion mask. The qualitative cases demonstrate both
successful corrections and remaining errors, while the calibration results
indicate that the performance improvements are accompanied by more reliable
confidence estimates. Finally, the efficiency analysis confirms that the
training-time use of masks and privileged supervision does not introduce
additional inference-time dependencies, preserving a lightweight and
mask-free deployment pipeline.

\section{Limitations and Future Work}
\label{sec:limitations_future_work}

Despite its strong internal and external performance, the proposed PLCRD
framework has several limitations. First, lesion masks are required during
training to construct the privileged lesion and context representations,
which may restrict applicability when reliable annotations are unavailable.
Although masks are not required during inference, preparing training-time
segmentations still introduces additional annotation cost. Second, the
evaluation is limited to HAM10000 and ISIC 2018, and therefore may not fully
capture the variability encountered across different clinical centers,
imaging devices, patient populations, and acquisition protocols. The current
study also relies on retrospective benchmark datasets and does not include a
prospective clinical validation or dermatologist reader study. In addition,
the interpretability results show that PLCRD produces highly precise
lesion-centered attention but relatively limited lesion coverage, indicating
that the model focuses on compact discriminative subregions rather than the
complete lesion extent. The remaining failure cases further demonstrate that
visually similar categories, such as melanoma, nevus, actinic keratosis, and
benign keratosis, can still be confused.

Future work will investigate weaker forms of privileged supervision, including
automatically generated pseudo-masks, point annotations, bounding boxes, and
self-supervised lesion localization, to reduce dependence on pixel-level
annotations. The framework should also be evaluated on additional
multi-institutional datasets and under prospective clinical settings to
assess robustness across domains. Incorporating patient metadata, lesion
history, anatomical location, and other clinical variables may further
improve diagnostic discrimination for visually ambiguous cases. Additional
directions include uncertainty-aware referral mechanisms, calibration under
distribution shift, class-specific relational modeling, and dermatologist
reader studies that evaluate whether PLCRD explanations improve human
decision-making. Finally, future work may explore more efficient unified
student architectures and stronger relation-learning objectives while
preserving the current mask-free deployment setting.

\section{Conclusion}
\label{sec:conclusion}

This work introduced PLCRD, a privileged lesion-context relational
distillation framework for mask-free skin-lesion classification. During
training, PLCRD transfers lesion-focused supervision and relational knowledge
between the lesion and its surrounding context from privileged mask-based
representations to an image-only student. At inference time, the privileged
teacher, lesion masks, and auxiliary distillation branches are removed,
allowing the final model to operate directly from dermoscopic images without
requiring a segmentation pipeline.
The experimental results demonstrate that PLCRD achieves strong internal
performance on HAM10000 and robust external generalization to ISIC 2018
without retraining. The proposed method obtains the highest balanced accuracy,
Macro-F1, and AUROC among the compared approaches, while also providing
improved calibration and competitive computational efficiency. The ablation
study confirms that the observed gains arise from the combined use of
privileged lesion supervision and lesion-context relational learning rather
than from direct ROI input, conventional knowledge distillation, or isolated
localization objectives. The interpretability analysis further shows that
PLCRD concentrates diagnostic evidence within compact lesion regions while
suppressing irrelevant background responses. Overall, PLCRD provides an
effective approach for transferring training-time lesion annotations into a
lightweight, calibrated, and mask-free classifier suitable for further
investigation in clinically oriented skin-lesion analysis.

\section*{Funding}
This research received no external funding.

\section*{Conflict of Interest}
The authors declare that they have no conflicts of interest.

\section*{Data Availability}
The datasets analyzed in this study are publicly available from
the HAM10000 and ISIC archives. The corresponding dataset references
and access information are provided in the manuscript. The implementation will be made available from the corresponding author upon reasonable request.

\section*{Ethics Approval} This study used publicly available, deidentified datasets and involved no direct interaction with human participants. Institutional ethics approval was therefore not required.

\section*{Consent to Participate} Not applicable. The study used publicly released, deidentified data.

\bibliographystyle{IEEEtran}
\bibliography{sample}

\end{document}